\documentclass[letterpaper]{article}
\usepackage[preprint]{aaai2027}
\copyrighttext{Preprint.}
\usepackage[hyphens]{url}
\usepackage{graphicx}
\usepackage{natbib}
\usepackage{caption}
\usepackage{booktabs}
\usepackage{array}
\usepackage{siunitx}
\usepackage{xcolor}
\usepackage{multirow}

\usepackage{amsmath}
\usepackage{amssymb}
\usepackage{mathtools}
\usepackage{amsthm}

\usepackage{enumitem}

\theoremstyle{plain}
\newtheorem{theorem}{Theorem}[section]

\theoremstyle{definition}
\newtheorem{definition}[theorem]{Definition}

\theoremstyle{remark}

\title{MASA: Multi-Agent Asynchronous Structural Adaptation for Efficient Heterogeneous Agent Systems}
\author{Fulin Lin\thanks{Equal contribution.\quad\textsuperscript{\textdagger}Corresponding author.}, Yuqing Huang\textsuperscript{*}, Zhaofen Wu, Ling Liu, Hongwei Wang\textsuperscript{\textdagger}}
\affiliations{Zhejiang University}

\begin{document}
\maketitle
\begin{abstract}
While Multi-Agent Systems (MAS) exhibit superior reasoning capabilities, their deployment is hindered by prohibitive inference latency and token costs.
Existing optimization methods typically decouple topology generation from model routing and rely on a static "generate-then-optimize" paradigm, failing to address the dynamic trade-off between task structure and model heterogeneity.
To address this, we propose \textsc{\textbf{Masa}} (\textbf{M}ulti-agent \textbf{A}synchronous \textbf{S}tructural \textbf{A}daptation), a training-free framework that performs online joint optimization of agent topology and backbone routing via a Stream-MCTS algorithm.
\textsc{Masa} iteratively refines the task stream using graph rewriting operators, while employing an asynchronous pipelining mechanism to effectively mask the search overhead.
This process is guided by a Max-Backup strategy, which balances throughput against agility to accurately navigate the combinatorial search space of heterogeneous agents.
Empirical results on complex reasoning benchmarks demonstrate that \textsc{Masa} achieves a superior Pareto frontier, significantly reducing end-to-end latency compared to static and decoupled baselines.
\end{abstract}

\section{Introduction}
\label{sec:intro}
The evolution of Large Language Models (LLMs) has catalyzed a paradigm shift from isolated conversational agents to sophisticated Multi-Agent Systems (MAS) \cite{gao2026surveyselfevolvingagentswhat,Tran2025MultiAgentCM}. Empowered by tool utilization and long-term memory \cite{NEURIPS2023_d842425e,park2023generativeagentsinteractivesimulacra}, agents have demonstrated remarkable proficiency across both general and specialized domains \cite{Tran2025MultiAgentCM,Luo2025LargeLM}. Compared to single-agent counterparts, MAS offers a higher performance ceiling through modular specialization, extended context windows, and inherent robustness via redundancy \cite{Liu2025AdvancesAC}. By decomposing complex queries into sub-tasks handled by specialized roles, MAS effectively mimics the "System 2" slow thinking process, enabling deeper reasoning and collaborative problem-solving \cite{Li2025FromS1,du2023improving}.

\begin{figure}[t]
  \begin{center}
    \centerline{\includegraphics[width=\columnwidth]{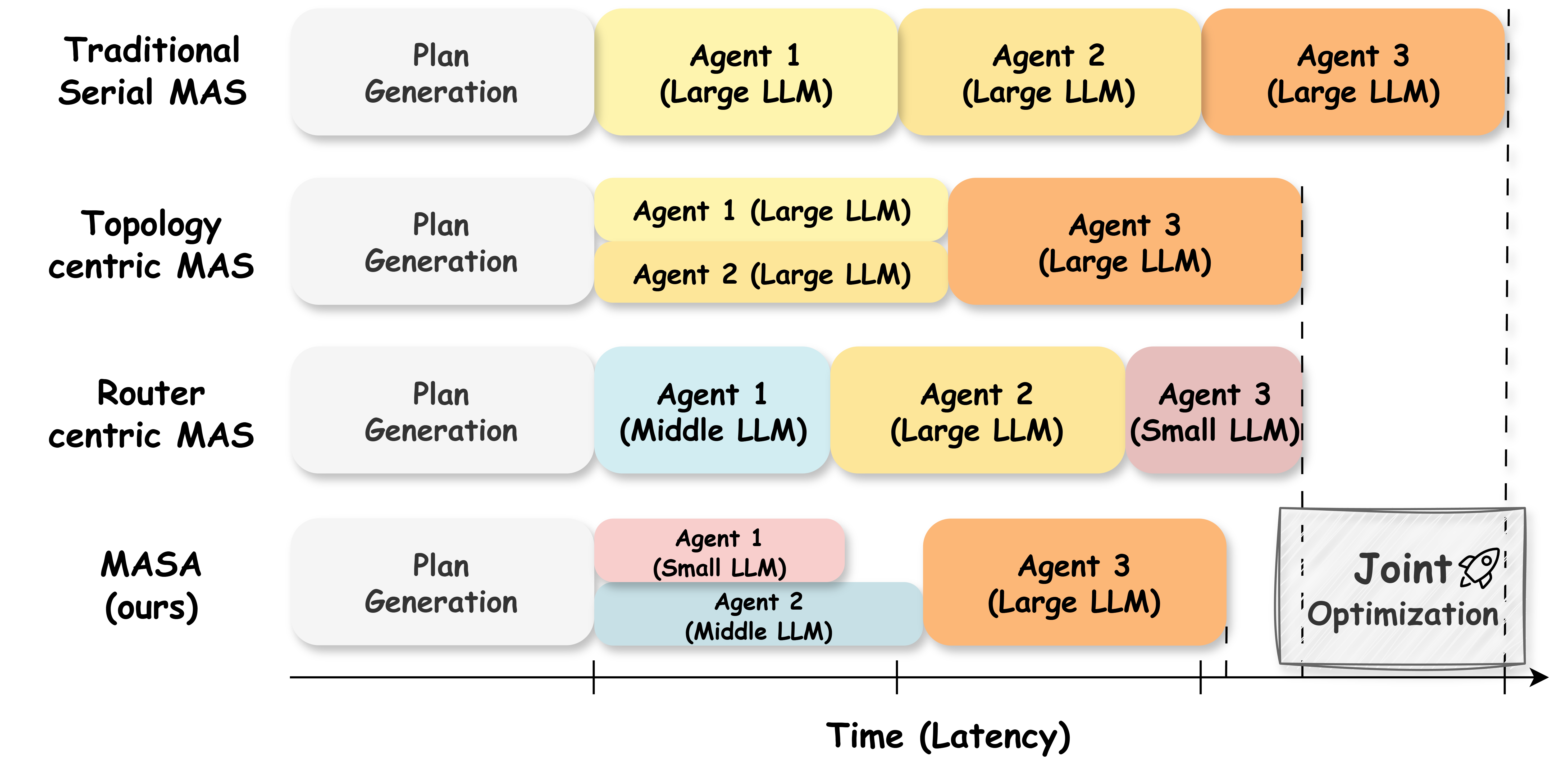}}
    \caption{\textbf{Comparison of different MAS paradigms.} Serial (ReAct-centric), topology-centric, and router-centric methods optimize reasoning traces, collaboration structures, and model selection separately, while \textsc{Masa} jointly optimizes routing and topology during runtime.}
    \label{fig:background}
  \end{center}
\end{figure}

However, this enhanced capability comes at a steep price. The collaborative nature of MAS necessitates frequent LLM invocations, intricate context maintenance, and multi-step reasoning chains \cite{ferrag2025llmreasoningautonomousai,zhang2025cut}, leading to prohibitive inference latency and exorbitant token costs \cite{li2025agentoriented,park2025surveyinferenceengineslarge}. These efficiency bottlenecks severely restrict the scalability and real-world deployment of MAS \cite{feng2026agentocrreimaginingagenthistory,wang2025agenttaxo}. Existing efforts to mitigate these issues typically fall into two isolated camps: \textit{Topology-centric} methods that optimize information flow structures to enhance parallelization \cite{zhang2025cut,zhang2025metaagent}, and \textit{Router-centric} methods that dynamically select model backbones to match task difficulty \cite{ong2025routellm,yue-etal-2025-masrouter}.

\begin{figure*}[t]
\centering
\includegraphics[width=1\linewidth]{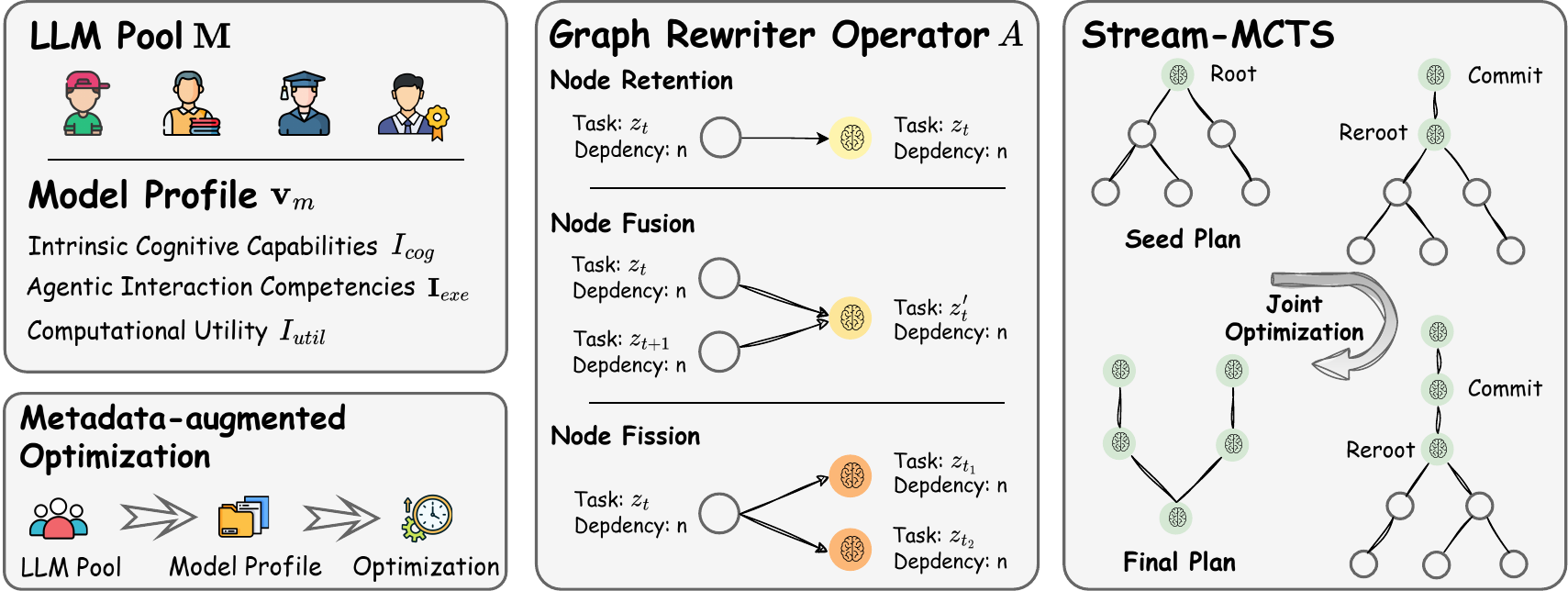}
\caption{\textbf{Overview of MASA.} The framework jointly optimizes model routing and collaboration topology through Stream-MCTS. The planner incrementally generates executable steps, while the optimization stream rewrites the collaboration graph and dynamically assigns models from a heterogeneous LLM pool using model profiles, metadata, and execution feedback.}
\label{fig:overview}
\end{figure*}

Despite their progress, these decoupled approaches face specific limitations.
First, the rapid evolution of LLMs renders training-based routers obsolete \cite{ong2025routellm}. Relying on neural routers necessitates frequent retraining to adapt to new backbones \cite{song2025a}. This dependency severely hinders the agility required to leverage state-of-the-art models instantly.
Second, static topological structures impose a rigid upper bound on efficiency. Standard architectures, such as ReAct \cite{Yao2022ReActSR} or Chain-of-Thought \cite{Wei2022ChainOT}, fail to adapt to the varying complexity of dynamic sub-tasks.

Third, the "overkill" phenomenon remains unaddressed. Static architectures often deploy the most powerful, yet slowest, models for even trivial sub-tasks \cite{fu-etal-2026-flare}. Conversely, while smaller models provide faster inference, they frequently lack the reasoning depth needed for complex nodes \cite{guan2025dynamicspeculativeagentplanning}.
Thus, current methods struggle to navigate the trade-off between the "System 1" speed of small models and the "System 2" depth of large models within a unified framework \cite{saha2025system}.

We argue that achieving the Pareto frontier of efficiency and performance requires a \textbf{Joint Optimization} of both the interaction topology and the backbone routing. As illustrated in Figure~\ref{fig:background}, compared to linear collaboration, a co-designed architecture can significantly compress task completion time by assigning cost-effective models to parallel sub-tasks.
To realize this vision, we must address three fundamental challenges:
(1) How to formalize the diverse topological variations using a finite set of graph operators?
(2) How to efficiently identify the optimal LLM backbone that matches sub-task difficulty without supervised training?
(3) How to perform this joint optimization in real-time without introducing significant computational overhead?

To address these challenges, we propose \textsc{\textbf{Masa}} (\textbf{M}ulti-agent \textbf{A}synchronous \textbf{S}tructural \textbf{A}daptation), a training-free framework for the online joint optimization of MAS topology and routing (see Figure~\ref{fig:overview}).
Building upon an initial \textit{Seed Plan}, \textsc{Masa} employs a \textbf{Stream-MCTS} algorithm to iteratively refine the agent collaboration graph using three graph rewriting operators: \textit{Fusion}, \textit{Fission}, and \textit{Retention}.
Crucially, to balance search determinism with execution flexibility, we decouple the "Macro-Optimization" of the system architecture from the "Micro-Execution" of individual agents. \textsc{Masa} optimizes the high-level collaboration structure and backbone allocation, while leaving atomic tool usage and reasoning steps to the individual agents. This design restricts the search space for fast convergence while preserving the general problem-solving capabilities of single agents.

Our core contributions are summarized as follows:
\begin{itemize}[leftmargin=*]
    \item \textbf{Problem Formulation of Joint Optimization:} We formally define the problem of joint topology and routing optimization for MAS. By modeling the system as a dynamic graph, we transform the efficiency challenge into a constrained multi-objective search problem over a discrete operator space.
    \item \textbf{Asynchronous Pipelining Framework:} We introduce a novel asynchronous architecture that parallelizes the \textit{Instruction}, \textit{Optimization}, and \textit{Execution} streams. This design achieves near-zero optimization overhead by masking the search latency behind the planner's generation and the executor's operation. Furthermore, our \textbf{Stream-MCTS} with a dynamic root and sliding window mechanism significantly accelerates the search process.
    \item \textbf{Comprehensive Evaluation:} We conduct extensive experiments across three complex reasoning benchmarks using heterogeneous LLM pools. Empirical results demonstrate that \textsc{Masa} significantly outperforms both topology-centric, router-centric, and hybrid baselines, achieving superior performance with reduced latency and cost.
\end{itemize}

\section{Related Work}

\paragraph{Efficiency.}  Current approaches to optimizing MAS efficiency typically balance performance against computational costs by relying on specific training objectives or heuristic constraints \cite{liu2025advanceschallengesfoundationagents}. In this landscape, topological methods  mitigate bottlenecks by excising redundant nodes \cite{wang2025agentdropoutdynamicagentelimination,zhang2025cut},while runtime frameworks such as SMAS \cite{lin2025stopwastingtokensefficient} and ACON \cite{kang2025aconoptimizingcontextcompression} employ adaptive filters to proactively intercept inefficient loops. Complementary speculative accelerators leverage look-ahead execution to parallelize reasoning processes \cite{ye2025speculativeactionslosslessframework,guan2025dynamicspeculativeagentplanning,huang2025reducinglatencyllmsearch}. Addressing the conflict between adaptation overheads and token wastage, our \textsc{Masa} framework introduces a cost-effective paradigm for flexible and precise optimization.

\paragraph{Dynamic Routing and Model Orchestration.}
Research in agent orchestration has shifted toward dynamic, learning-based routing. ToolOrchestra \cite{toolorchestra} and ScoreFlow \cite{wang2025scoreflowmasteringllmagent} optimize model selection through reinforcement learning, while AgentRouter \cite{zhang2025agentrouterknowledgegraphguidedllmrouter}, GraphRouter \cite{feng2025graphroutergraphbasedrouterllm}, and AFlow \cite{zhang2025aflow} use graph modeling or search to navigate routing and workflow spaces. Difficulty-aware methods such as DAAO \cite{su2025difficultyawareagenticorchestrationqueryspecific} adapt routing to query complexity, and MasRouter \cite{yue-etal-2025-masrouter} jointly determines the collaboration mode, agent roles, and LLM assignments. GraphPlanner \cite{feng2026graphplanner} jointly selects agent roles and LLM backbones to generate query-specific workflows, using heterogeneous graph memories and a PPO-trained policy. \textsc{Masa} likewise coordinates workflow structure and backbone selection, but uses training-free online search over explicit topology-rewriting and model-assignment decisions. Its asynchronous streaming pipeline further overlaps planning, optimization, and execution.

\paragraph{Dynamic MAS Topology.}
Constructing effective MAS communication topologies often requires searching over structural spaces or adapting from interaction history. Prior work automates this process through MCTS-based reasoning or workflow search, such as LATS \cite{zhou2024language} and AFlow \cite{zhang2025aflow}, differentiable or modular architecture search, such as MaAS \cite{zhang2025agentic-supernet} and AgentSquare \cite{shangagentsquare}, and runtime orchestration mechanisms, such as Puppeteer \cite{dang2025multi} and EvoRoute \cite{zhang2026evoroute}. Generative methods further synthesize task-specific agents or graph templates, including AutoAgents \cite{chen2024autoagents} and AMAS \cite{leong2025amas}. However, these methods mostly operate at \textit{step-scale}, focusing on shallow reasoning structures or restricted collaboration patterns, often with homogeneous backbones. This limits their scalability on complex long-horizon tasks; in contrast, our method targets heterogeneous \textit{MAS-scale} topology generation and routing.

\begin{figure*}[t]
\centering
\includegraphics[width=1\linewidth]{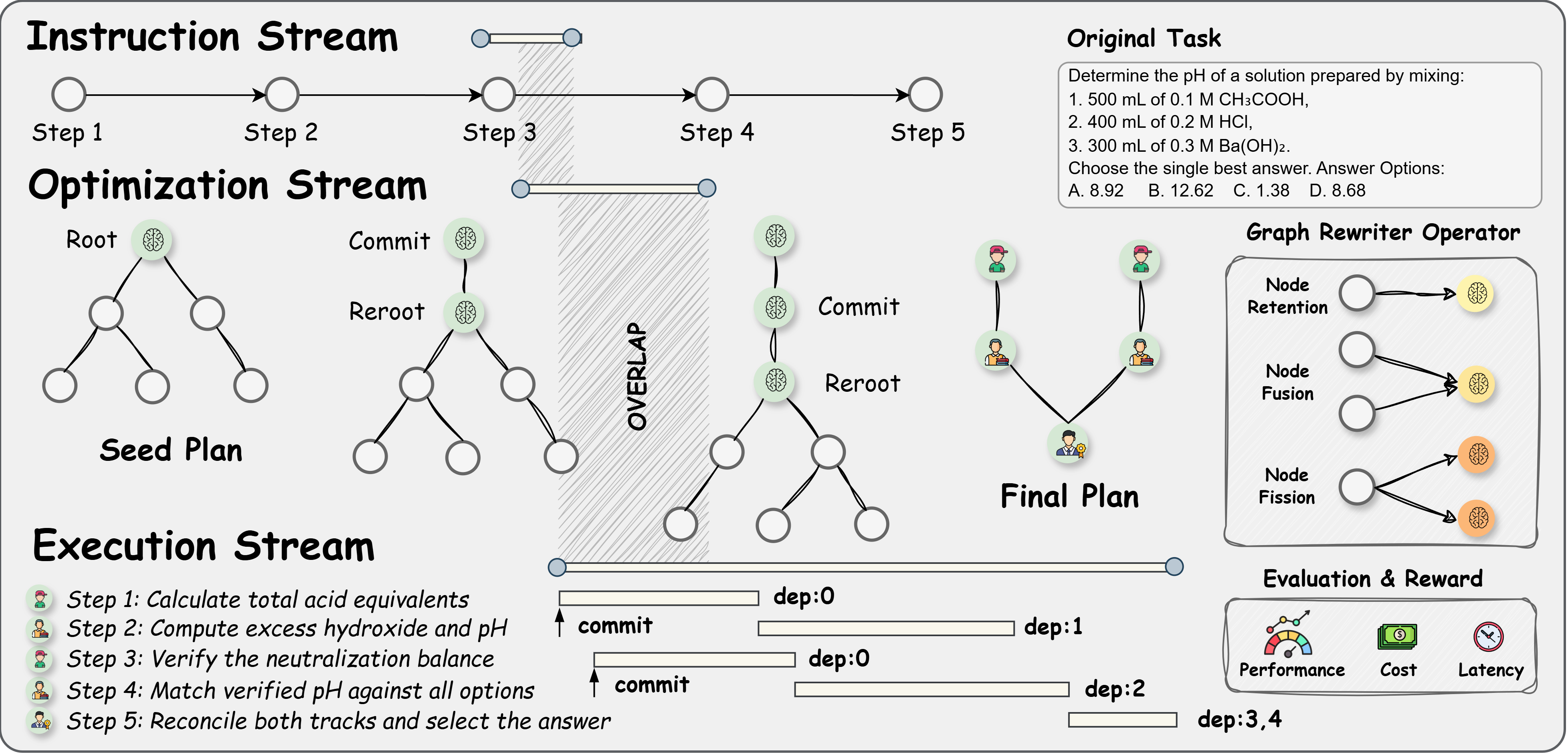}
\caption{\textbf{Asynchronous execution pipeline of MASA.} The Instruction Stream continuously generates executable steps, the Optimization Stream incrementally rewrites and routes the collaboration graph through Stream-MCTS, and the Execution Stream starts node execution before the full plan is finalized, enabling overlap between planning, optimization, and execution to reduce end-to-end latency.}
\label{fig:framework}
\end{figure*}

\section{Preliminary}
\label{sec:preliminary}

\subsection{Problem Formulation}

\paragraph{Graph Construction and Search Space.}
We first formalize the fundamental building block of our streaming multi-agent system, namely the configurable agent node.

\begin{definition}[\textbf{Configurable Agent Node}]
\label{def:agent_node}
A configurable agent node $v$ is a tuple representing an atomic execution unit, characterized by its instruction context and the assigned computational backbone. For the $i$-th node:
\begin{equation}
\label{eq:node_def}
\begin{gathered}
v_i = \langle \mathcal{I}_i, m_k \rangle,
\; \mathcal{I}_i \in \mathcal{D}, \; m_k \in \mathcal{M},
\end{gathered}
\end{equation}
where $\mathcal{D}$ denotes the semantic space of natural language instructions (prompts), and $\mathcal{M} = \{m_1, \dots, m_K\}$ represents the finite pool of heterogeneous LLM backbones. The assignment $m_k$ constitutes the routing decision.
\end{definition}

Based on this definition, we model the Multi-Agent System (MAS) as a directed acyclic graph (DAG). The structural space is defined as:
\begin{equation}
\label{eq:graph_def}
\begin{gathered}
\mathcal{G} = (\mathcal{V}, \mathcal{E}), \; \mathcal{V} = \{v_1, \dots, v_n\}, \; \mathcal{E} \subseteq \mathcal{V} \times \mathcal{V}, \\
\end{gathered}
\end{equation}
where $n = |\mathcal{V}|$ is the dynamic number of agents. Consequently, the joint search space $\Omega$ for topology and routing grows exponentially with the graph scale. Theoretically, the unconstrained search space can be approximated as:
\begin{equation}
\label{eq:search_space}
\Omega \approx \sum_{n=1}^{N_{max}} \underbrace{(\mathcal{D} \times \mathcal{M})^n}_{\text{Routing Space}} \times \underbrace{2^{n^2}}_{\text{Topology Space}},
\end{equation}
where $N_{max}$ is the maximum allowable steps. Since $\mathcal{D}$ is theoretically infinite, we reduce $\mathcal{D}$ to a finite set of system prompts to render the problem tractable for MCTS.

\paragraph{Sequential Graph Generation.}
The construction of the optimal execution graph is formulated as a sequential decision-making process. We define the dynamics of this process using graph rewriting operators.

\begin{definition}[\textbf{Sequential Decision Dynamics}]
Let $\mathcal{A}$ be the set of abstract graph rewriting operators. The graph generation follows a sequential decision process governed by a policy $\pi$ and a transition function $\mathcal{T}$. At step $t$, the policy determines the action based on the current state, which triggers the LLM-driven transition:
\begin{equation}
a_t = \pi(\mathcal{S}_t), \quad \mathcal{S}_{t+1} = \mathcal{T}(\mathcal{S}_t, a_t).
\end{equation}
Here, $\pi$ represents the optimization agent selecting operators, while $\mathcal{T}$ represents the semantic execution engine generating the actual node content and updating the graph topology.
\end{definition}

\paragraph{Joint Optimization Objective.}
Given an initial user query $q$, our objective is to identify the optimal action sequence that constructs the target graph $\mathcal{G}^*$ maximizing a multi-objective utility function. We formulate this as a discrete optimization problem:
\begin{equation}
\label{eq:objective}
\begin{aligned}
\mathcal{G}^* &= \underset{\mathcal{G} \in \Omega'}{\arg\max} \; \mathcal{J}(\mathcal{G}) \\
&= \underset{\mathcal{G}}{\arg\max} \; \Big[ Q(\mathcal{G}; q) - \gamma \cdot C(\mathcal{G}) \\
&\quad - \delta \cdot L(\mathcal{G}) \Big],
\end{aligned}
\end{equation}
where $\Omega'$ denotes the subspace of graphs constructible via sequences of operators in $\mathcal{A}$. $Q(\cdot)$, $C(\cdot)$, and $L(\cdot)$ represent the semantic quality score, token consumption cost, and end-to-end latency, respectively. $\gamma, \delta \in \mathbb{R}^+$ are Lagrange multipliers governing the trade-off between performance and efficiency.

\subsection{Agent Heterogeneity Modeling}
\label{sec:agent_profile}

Effective routing within the joint optimization framework requires a structured representation of the heterogeneous capabilities of LLMs. Rather than treating models as opaque identifiers, we map each backbone $m \in \mathcal{M}$ to a multidimensional profile vector.

\paragraph{Multidimensional Capability Profiling.}
We define a profiling function $\Phi: \mathcal{M} \to \mathbb{R}^d$ that projects each discrete model into a continuous capability space. The profile vector $\mathbf{v}_m$ is decomposed into three orthogonal dimensions:
\begin{equation}
\label{eq:profile_vec}
\mathbf{v}_m = \Phi(m) = \left[ I_{\text{cog}}(m), \; \mathbf{I}_{\text{exe}}(m), \; I_{\text{util}}(m) \right]^\top,
\end{equation}
where
$I_{\text{cog}} \in \mathbb{R}$ quantifies \textit{Intrinsic Cognitive Capabilities}, synthesizing deep reasoning and domain knowledge benchmarks. $\mathbf{I}_{\text{exe}} \in \mathbb{R}^3$ represents \textit{Agentic Interaction Competencies}, a vector capturing tool-use precision, format compliance, and error resilience. $I_{\text{util}} \in \mathbb{R}$ denotes the \textit{Computational Utility}, a dual-pathway metric balancing throughput efficiency against latency constraints.

Detailed formulations and data sources for these metrics are provided in Appendix.

\paragraph{Noisy Priors and Robustness.}
Crucially, the profile $\mathbf{v}_m$ is derived from public third-party benchmarks, which act as \textit{heuristic priors} rather than ground-truth guarantees. We model the observed capability $\hat{\mathbf{v}}_m$ as an approximation of the true capability $\mathbf{v}^*_m$ subject to epistemic uncertainty:
\begin{equation}
\hat{\mathbf{v}}_m = \mathbf{v}^*_m + \epsilon, \; \epsilon \sim \mathcal{N}(\mathbf{0}, \Sigma), \; \Sigma \in \mathbb{R}^{d \times d},
\end{equation}
where $\epsilon \in \mathbb{R}^d$ represents the noise introduced by benchmark contamination or distribution shifts.
Consequently, simply maximizing a static utility function based on $\hat{\mathbf{v}}_m$ is prone to sub-optimal routing.
This necessitates the use of our MCTS-based framework. By treating $\hat{\mathbf{v}}_m$ as heuristic guidance for the selection policy, the search algorithm dynamically validates these priors through rollout simulations, thereby achieving robustness against the inherent unreliability of external evaluations.

\section{Methodology}
\label{sec:methodology}

\subsection{Asynchronous Pipelining Framework}
\label{sub:framework}

To resolve the joint optimization problem defined in Preliminary, we propose \textsc{\textbf{Masa}} (Multi-agent Asynchronous Structural Adaptation). Unlike traditional "Generate-then-Optimize" paradigms that block execution, \textsc{Masa} utilizes a \textbf{training-free asynchronous pipelining mechanism}. As shown in Figure~\ref{fig:framework}, the system operates across three concurrent streams:

\begin{itemize}[leftmargin=*]
    \item \textbf{Instruction Stream ($\mathcal{S}_{\text{ins}}$):} Driven by a Planner LLM, this stream generates the seed plan $\mathcal{Z} = \{z_1, z_2, \dots\}$ sequentially. Each $z_t$ acts as a raw instructional unit containing dependencies but lacking model allocation.
    \item \textbf{Optimization Stream ($\mathcal{S}_{\text{opt}}$):} Governed by our \textbf{Stream-MCTS} algorithm, this stream performs online receding horizon search. It continuously monitors $\mathcal{S}_{\text{ins}}$ and transforms raw steps $z$ into configured agent nodes $v$.
    \item \textbf{Execution Stream ($\mathcal{S}_{\text{exe}}$):} This stream consumes the \textit{committed} nodes from $\mathcal{S}_{\text{opt}}$ and instantly dispatches them to the assigned backbones in $\mathcal{M}$.
\end{itemize}

A critical advantage of this design is \textbf{Latency Masking}: most of the optimization overhead is effectively hidden as $\mathcal{S}_{\text{opt}}$ operates in parallel with the ongoing generation of $\mathcal{S}_{\text{ins}}$ and the physical execution in $\mathcal{S}_{\text{exe}}$.

\subsection{Graph Rewriting Operators}
\label{sub:operators}

We instantiate the abstract action space $\mathcal{A}$ with three specific operators. These operators function as the transition logic $\mathcal{T}$, transforming the local instruction window $\mathcal{W}_t = (z_t, z_{t+1})$ into configured agent nodes. Note that these operators modify not only the topological structure but also the \textit{instruction context} $\mathcal{I}$ to maintain semantic consistency.

\paragraph{Node Retention (Direct Routing).}
This operator maintains the atomic structure of the seed instruction $z_t$ and focuses purely on selecting the optimal backbone. It serves as the default transition:
\begin{equation}
a_{\text{ret}}: z_t \xrightarrow{\mathcal{T}} v_t = \langle \mathcal{I}_t, m^* \rangle, \quad m^* \in \mathcal{M}.
\end{equation}

\paragraph{Node Fusion (Topology Compression).}
To leverage the long-context capabilities of stronger models, this operator merges $z_t$ with its successor $z_{t+1}$. The transition logic \textit{concatenates} their instructions into a unified context, reducing graph depth:
\begin{equation}
a_{\text{fuse}}: (z_t, z_{t+1}) \xrightarrow{\mathcal{T}} v_{\text{fused}} = \langle \textsc{Merge}(\mathcal{I}_t, \mathcal{I}_{t+1}), m^* \rangle.
\end{equation}

\paragraph{Node Fission (Topology Expansion).}
For complex tasks exceeding single-model capabilities, this operator decomposes $z_t$ into sub-tasks. The transition \textit{splits} the original instruction $\mathcal{I}_t$ into semantically distinct sub-instructions $\mathcal{I}_a, \mathcal{I}_b$ to execute:
\begin{equation}
\begin{aligned}
a_{\text{fis}}: z_t &\xrightarrow{\mathcal{T}} \{v_a, v_b\}, \\
\text{where } v_a &= \langle \mathcal{I}_a, m_j \rangle, \; v_b = \langle \mathcal{I}_b, m_k \rangle.
\end{aligned}
\end{equation}

\subsection{Stream-MCTS: Online Receding Horizon Search}
\label{sub:stream_mcts}

The core of $\mathcal{S}_{\text{opt}}$ is \textbf{Stream-MCTS}, adapted for long-horizon streaming tasks. It employs a \textbf{Sliding Window} of size $w=2$ to limit the branching factor and focuses computational resources on the immediate decision frontier.

\paragraph{Tree Policy and Expansion.}
We traverse the search tree using the \textit{UCT} criterion to balance exploration and exploitation. Upon reaching a leaf node, we employ a lightweight proxy policy $\pi_{\text{light}}$ to generate potential actions, LLM profile vector $a \in \mathcal{A}$ based on the current context, the profile vector $\mathbf{v}_m$ and available models $\mathcal{M}$:
\begin{equation}
\{a^{(1)}, \dots, a^{(k)}\} \sim \pi_{\text{light}}(\mathcal{A} \mid \mathcal{W}_t, \mathcal{M}, \mathbf{v}_m).
\end{equation}
This allows for rapid topology enumeration without invoking heavy backbones.

\paragraph{Predictive Evaluation via LLM Proxy.}
Unlike traditional methods that require full execution to estimate value, we propose a cost-effective \textbf{Predictive Evaluation} mechanism. We utilize the lightweight policy $\pi_{\text{light}}$ to assess the compatibility between the assigned model $m_k$ and the specific step instruction $\mathcal{I}_t$, utilizing the model's profile $\mathbf{v}_m$ as context:
\begin{equation}
q_m, d \sim \pi_{\text{light}}(q, d \mid \mathcal{I}_t, m_k, \mathbf{v}_m),
\end{equation}
where $q_m \in [0, 1]$ represents the \textbf{Compatibility Score}, and $d \in \{\text{Low}, \text{Mid}, \text{High}\}$ denotes the estimated \textbf{Subtask Difficulty}.

The difficulty $d$ serves as a proxy to retrieve expected token counts from a pre-calibrated look-up table. We then compute the estimated cost $\hat{C}$ and latency $\hat{L}$:
\begin{align}
\hat{C} &= T_{\text{in}}(d) \cdot P_{\text{in}}^{(m)} + T_{\text{out}}(d) \cdot P_{\text{out}}^{(m)}, \\
\hat{L} &= \text{TTFT}^{(m)} + c \cdot \frac{T_{\text{out}}(d)}{\text{Speed}^{(m)}},
\end{align}
where $T_{\text{in/out}}(d)$ are the expected token counts for difficulty $d$, and $P_{\text{in/out}}^{(m)}$ are the unit prices for model $m$. In the latency formula, $\text{TTFT}$ and $\text{Speed}$ are retrieved from the profile $\mathbf{v}_m$, and $c$ is a penalty coefficient for reasoning-intensive models to account for non-linear generation delays.

\paragraph{Scalarized Reward and Simultaneous Update.}
To accelerate convergence, we incorporate a progress signal $\mathcal{P}_{\text{prog}}$ (current step depth / predicted total steps). The final reward $R(v)$ is formulated as:
\begin{equation}
\label{eq:reward}
R(v) = \alpha \cdot \mathcal{P}_{\text{prog}} + \beta \cdot q_m - \gamma \cdot \hat{C} - \delta \cdot \hat{L},
\end{equation}
where $\mathcal{P}_{\text{prog}} = n_{\text{step}}/n_{\text{total}}$ quantifies progress, and $\{\alpha, \beta, \gamma, \delta\}$ are trade-off hyperparameters. To minimize latency, we implement parallel evaluation, where sibling nodes generated during expansion are scored concurrently.

\paragraph{Max-Backup and Asynchronous Commit.}
We employ a \textbf{Max-Backup} strategy: $V(n) \leftarrow \max_{n' \in \text{children}(n)} V(n')$, accelerating convergence toward high-value trajectories. To handle the continuous stream, a \textbf{Commit Mechanism} solidifies the root decision when the visit count $N_{root}$ exceeds a threshold $N_{th}$. The committed node $v^*$ is pushed to $\mathcal{S}_{\text{exe}}$, and the window slides forward as well.

\section{Experiments}

\subsection{Experimental Setup}

\begin{table*}[t]
\centering
\small
\setlength{\tabcolsep}{3pt}
\resizebox{\textwidth}{!}{\begin{tabular}{@{} l l
  *{1}{S[table-format=3.2, detect-weight=true]}
  *{1}{S[table-format=1.4]}
  *{1}{S[table-format=3.2]}
  *{1}{S[table-format=2.2, detect-weight=true]}
  *{1}{S[table-format=2.4]}
  *{1}{S[table-format=3.2]}
  *{2}{S[table-format=2.2, detect-weight=true]}
  *{1}{S[table-format=2.4]}
  *{1}{S[table-format=4.2]}
  @{}}
\toprule
Type & Baseline
& \multicolumn{3}{c}{AIME 25}
& \multicolumn{3}{c}{GPQA}
& \multicolumn{4}{c}{SciCode} \\
\cmidrule(lr){3-5}
\cmidrule(lr){6-8}
\cmidrule(lr){9-12}
&
& \multicolumn{1}{c}{pass@3 (\%)}
& \multicolumn{1}{c}{Cost (\$)}
& \multicolumn{1}{c}{Latency (s)}
& \multicolumn{1}{c}{pass@3 (\%)}
& \multicolumn{1}{c}{Cost (\$)}
& \multicolumn{1}{c}{Latency (s)}
& \multicolumn{1}{c}{Main (\%)}
& \multicolumn{1}{c}{Sub (\%)}
& \multicolumn{1}{c}{Cost (\$)}
& \multicolumn{1}{c}{Latency (s)} \\
\midrule

\multirow{4}{*}{Vanilla}
& Qwen3-8B
& 29.17 & 0.0647 & 42.69
& 40.25 & 0.1113 & 6.90
& 3.08 & 24.40 & 0.8382 & 58.41 \\

& Qwen3-14B
& 33.33 & 0.1857 & 58.98
& 65.41 & 0.0471 & 0.62
& 3.08 & 29.90 & 1.5402 & 65.13 \\

& Qwen3-30B-a3b-instruct
& 45.83 & 0.1890 & 42.74
& 55.35 & 0.2564 & 6.38
& 4.62 & 32.99 & 1.2279 & 50.01 \\

& Qwen3-30B-a3b-thinking
& 91.67 & 2.1285 & 197.82
& 78.62 & 3.9840 & 55.53
& 9.23 & 32.99 & 10.4943 & 254.51 \\

\midrule

Router (model list)
& MASRouter
& 70.83 & 1.6919 & 123.16
& 81.76 & 5.6694 & 60.45
& 1.54 & 35.05 & 41.6306 & 1049.59 \\

\midrule

Topology (instruct)
& AFlow
& 70.83 & 9.9441 & 71.12
& 80.50 & 7.0540 & 43.25
& 6.15 & 34.02 & 21.3361 & 344.43 \\

\midrule

Hybrid (model list)
& GraphPlanner
& 62.50 & 0.9781 & 54.91
& 76.10 & 6.0015 & 49.01
& 9.23 & 36.08 & 10.9724 & 260.33 \\

\midrule

MAS (thinking)
& OAgents
& 83.33 & 3.4359 & 256.77
& 81.76 & 19.5989 & 178.36
& 9.23 & 37.11 & 38.6758 & 542.94 \\

\midrule

Ours
& \textbf{MASA}
& \bfseries 100.00 & 0.4390 & 44.76
& \bfseries 86.79 & 9.3652 & 77.49
& \bfseries 10.77 & 36.43 & 7.5308 & 99.05 \\

\bottomrule
\end{tabular}}
\caption{\textbf{Main results} (pass@3) on AIME25, GPQA, and SciCode. We compare \textsc{Masa} against vanilla single-model inference, router-only methods, topology-only methods, hybrid methods, and general MAS baselines under performance, cost, and latency metrics. \textsc{Masa} consistently improves the Pareto frontier, achieving stronger reasoning performance while reducing the cost and latency overhead typically introduced by large multi-agent systems.}
\label{tab:main_results}
\end{table*}

We conduct a rigorous empirical evaluation to assess the efficacy of \textsc{Masa} across diverse reasoning domains. This section details our experimental setup, comprising the selected datasets, comparative baselines, and evaluation metrics. Extensive implementation details are provided in Appendix.

\paragraph{Datasets.} We evaluate on three challenging benchmarks: AIME 2025 \cite{aime25} (advanced math), SciCode \cite{tian2024scicode} (hierarchical scientific coding), and GPQA-Diamond \cite{rein2024gpqa} (expert scientific synthesis). For each benchmark, we use a fixed, seeded train--test split of approximately 1:4. All trainable baselines use the same training subset. In the main comparison, \textsc{Masa} and all baselines run on the same held-out problems without access to test labels. The SciCode test set contains 65 main problems and 291 subtasks.

\paragraph{Baselines.} We compare \textsc{Masa} with four representative agentic systems that isolate the design dimensions of interest: AFlow \cite{zhang2025aflow} optimizes workflow topology, MasRouter \cite{yue-etal-2025-masrouter} improves model routing for MAS, GraphPlanner \cite{feng2026graphplanner} jointly optimizes agent roles and model routing to construct query-specific topologies, and OAgents \cite{zhu-etal-2025-oagents} represents a general-purpose multi-agent system; vanilla LLM inference provides the single-model reference.

\subsection{Main Results}
\label{sec:main-results}

\begin{figure}[h]
\centering
\includegraphics[width=1\linewidth]{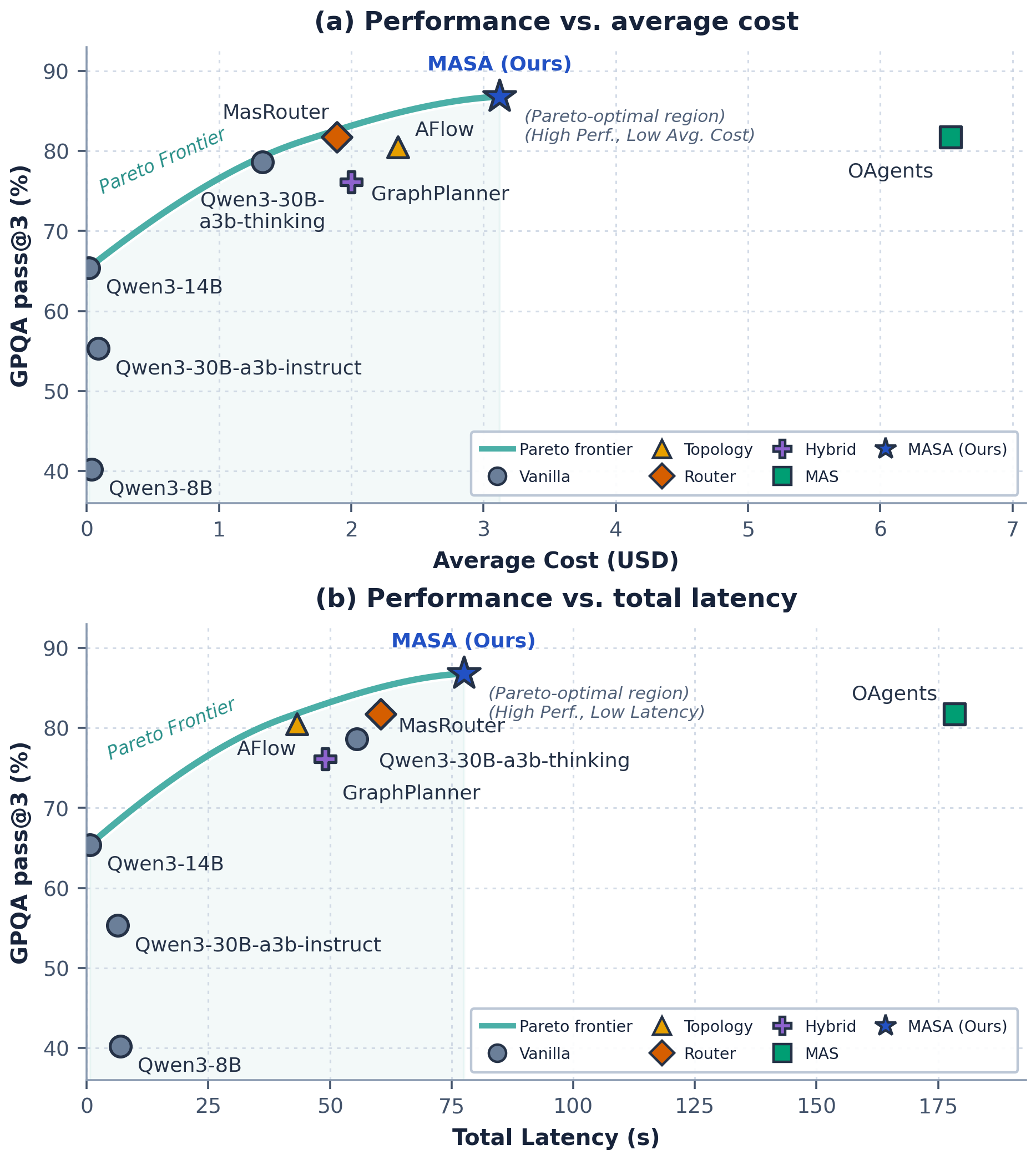}
\caption{\textbf{Pareto frontier on GPQA.} \textsc{Masa} shifts the frontier toward higher accuracy compared with router-only, topology-only, hybrid, and general MAS baselines.}
\label{fig:pareto}
\end{figure}

Table~\ref{tab:main_results} reports the pass@3 comparison across mathematical reasoning, scientific QA, and scientific coding. Overall, \textsc{Masa} consistently occupies the high-performance region of the accuracy--cost--latency Pareto frontier across all three benchmarks (see Figure~\ref{fig:pareto}). The advantage is most evident on reasoning-intensive tasks, where stronger single-model and multi-agent baselines typically improve performance at substantially higher inference cost or latency.

On \textbf{AIME25}, \textsc{Masa} achieves 100.00\% pass@3, exceeding the strongest baseline by 8.33 points while being 4.8$\times$ cheaper and 4.4$\times$ faster. On \textbf{GPQA}, it reaches the best pass@3 score of 86.79\%, improving over MASRouter and OAgents by 5.03 points, while substantially reducing cost and latency relative to OAgents. On \textbf{SciCode}, \textsc{Masa} obtains the highest main-task score of 10.77\% and a subtask score of 36.43\% (106/291), two solved subtasks behind OAgents (108/291, a gap of 0.69 percentage points), while reducing latency to 99.05s. Compared with GraphPlanner, it improves main-task performance by 1.54 points while being 1.46$\times$ cheaper and 2.63$\times$ faster. These results demonstrate that the instruction--optimization--execution pipeline consistently improves reasoning quality while maintaining practical cost and latency across heterogeneous task domains.

\begin{table}[h]
\centering
\small
\setlength{\tabcolsep}{3pt}
\resizebox{\columnwidth}{!}{\begin{tabular}{lcccc}
\toprule
\textbf{Method / Model} &
\textbf{Main (\%)} &
\textbf{Sub (\%)} &
\textbf{Cost (\$)} &
\textbf{Latency (s)} \\
\midrule
DeepSeek V4 Flash
& 13.85 & 41.92 & 0.2303 & 47.93 \\
Claude Sonnet 4.6
& 13.85 & 43.30 & 17.8963 & 122.44 \\
GPT-5.6 Terra
& 16.92 & 46.05 & 5.8190 & 85.51 \\
Qwen3.7 Max
& 18.46 & 52.23 & 15.2232 & 496.44 \\
\textbf{MASA (mixed pool)}
& \textbf{27.69} & \textbf{61.17} & 6.7273 & 151.87 \\
\bottomrule
\end{tabular}}
\caption{\textbf{Cross-family SciCode results.} MASA uses a mixed pool of GPT, Qwen, DeepSeek, and Claude models and achieves the strongest Main and Sub performance with competitive cost and latency.}
\label{tab:cross_family}
\end{table}

\paragraph{Cross-Family Generalization.}
We further evaluate \textsc{Masa} on SciCode using a pool spanning GPT, Qwen, DeepSeek, and Claude, including recently released backbones. As shown in Table~\ref{tab:cross_family}, \textsc{Masa} improves the strongest single-model Main and Sub scores by 9.23 and 8.93 percentage points, respectively. On the 291 subtasks, the mixed pool solves 178, compared with 152 for Qwen3.7 Max. These results demonstrate that its profile-based routing and graph rewriting transfer across model families and can incorporate new backbones without retraining.

\begin{table}[h]
\centering
\small
\begin{tabular}{lccc}
\toprule
\textbf{Configuration} &
\textbf{Avg.@3 (\%)} &
\textbf{Cost (\$)} &
\textbf{Latency (s)} \\
\midrule
\textbf{Full MASA} &
\textbf{90.28 $\pm$ 4.81} &
\textbf{0.44} &
\textbf{44.76} \\
w/o Sliding Window &
90.28 $\pm$ 8.67 &
0.53 &
45.54 \\
w/o Max Backup &
86.11 $\pm$ 4.81 &
0.51 &
47.25 \\
w/o Prediction &
88.89 $\pm$ 6.36 &
0.51 &
42.04 \\
\bottomrule
\end{tabular}
\caption{\textbf{Ablation on AIME25.} Each Stream-MCTS component is removed separately and evaluated by Avg@3 accuracy, cost, and latency.}
\label{tab:ablation}
\end{table}

\paragraph{Ablation Analysis.}
Table~\ref{tab:ablation} isolates the roles of three Stream-MCTS components. Removing the Sliding Window preserves mean accuracy but increases its variance, cost by 20.5\%, and latency by 1.7\%, showing that early commitment mainly improves efficiency and stability. Removing Max Backup causes the accuracy drop of 4.17 points and increases both cost and latency, indicating that it helps concentrate search on high-value branches. Removing Subtask Prediction lowers accuracy by 1.39 points and increases cost by 15.9\%, although latency decreases by 6.1\%, reflecting a trade-off from its additional planning overhead. Overall, the full method provides the most balanced performance.

\begin{table}[t]
\centering
\small
\setlength{\tabcolsep}{3.5pt}
\resizebox{\columnwidth}{!}{\begin{tabular}{ccccc}
\toprule
$\mathbf{w}$ &
\textbf{Avg.@3 (\%)} &
\textbf{Cost (\$)} &
\textbf{Total Lat. (s)} &
\textbf{Plan Lat. (s)} \\
\midrule
1 & 88.89 $\pm$ 6.36 & 0.46 & 43.74 & 13.32 \\
2 & 90.28 $\pm$ 4.81 & \textbf{0.44} & 44.76 & 14.34 \\
4 & \textbf{91.67 $\pm$ 0.00} & 0.53 & 50.02 & 16.28 \\
8 & 90.28 $\pm$ 6.36 & 0.51 & 47.60 & 16.74 \\
\bottomrule
\end{tabular}}
\caption{\textbf{Sliding window sensitivity on AIME25.} Increasing $w$ improves search lookahead but raises planning overhead, with $w=2$ providing the best cost-efficiency trade-off.}
\label{tab:window_sensitivity}
\end{table}

\paragraph{Sliding Window Size Sensitivity.}
The sliding window size $w$ controls the trade-off between search lookahead and online optimization overhead. Table~\ref{tab:window_sensitivity} shows that increasing $w$ generally raises planning latency, while total latency remains non-monotonic because improved topology and routing decisions can reduce downstream execution time. Although $w=4$ achieves the highest Avg@3, $w=2$ provides the lowest cost with near-lowest latency, motivating its use as the default Pareto-efficient setting.

\section{Discussion and Conclusion}
\label{sec:discussion_conclusion}

\paragraph{Discussion.}
Our results suggest that efficient heterogeneous MAS depends more on selecting the appropriate model for each subtask than on uniformly using the largest or most expensive backbone.
The cross-family evaluation shows that allocating complementary backbones to different subtasks can outperform every individual backbone, while the ablations indicate that topology adaptation and predictive routing contribute in different ways.
Sliding Window and Max Backup mainly reduce redundant execution and coordination overhead, whereas Subtask Prediction improves model--task matching at a modest planning cost. These findings support a task-dependent view of model capability, where reasoning strength, tool use, latency, and reliability should be considered jointly rather than reduced to model scale alone.

Training-free routing is particularly valuable as model pools evolve rapidly. Unlike learned routers that require retraining when new backbones are introduced, \textsc{Masa} can incorporate recently released models through lightweight profiling and test-time search. Its asynchronous pipeline further reduces end-to-end latency by overlapping graph construction, routing, and execution rather than treating them as blocking stages. However, fine-grained model switching may reduce KV-cache reuse and increase context-transfer overhead in persistent production workloads. The current framework also relies on approximate model profiles, lightweight predictive policies, and verifier-aware finalization, whose reliability may vary under distribution shift. Serving-aware routing, coarser agent-level decisions, better uncertainty calibration, and automated profile construction remain important directions for deployment at scale.

\paragraph{Conclusion.}
We present \textsc{Masa}, a training-free framework that jointly optimizes collaboration topology and model routing during inference. Stream-MCTS incrementally rewrites the execution graph, while the asynchronous pipeline overlaps search with downstream execution. Experiments on mathematical reasoning, scientific QA, and coding show that \textsc{Masa} consistently reaches the high-performance region of the accuracy--cost--latency pareto frontier. The ablation, sensitivity, and cross-family results further demonstrate the complementary roles of online graph optimization, predictive routing, and asynchronous execution. Future work will explore serving-aware routing, stronger uncertainty calibration, and automated adaptation to larger and continually evolving heterogeneous model pools.

\bibliography{citation}

\clearpage
\setcounter{secnumdepth}{2}
\appendix
\section{Experimental Setup}
\label{app:experimental}

\subsection{Datasets}

\label{sec:datasets}

Here, we provide a detailed introduction to the datasets used in this paper.
\begin{itemize}[leftmargin=*, nosep, itemsep=2pt, parsep=0pt, partopsep=0pt]

\item\textbf{AIME 2025} \cite{aime25} contains 30 problems from the 2025 American Invitational Mathematics Examination, covering algebra, geometry, number theory, and combinatorics. We use 6 problems for training and 24 for testing.

\item \textbf{SciCode} \cite{tian2024scicode} is a challenging, scientist-curated benchmark designed to evaluate models on realistic scientific research workflows. Spanning diverse natural science domains (e.g., physics, mathematics, biology, and chemistry), it comprises 80 complex main problems decomposed into 338 hierarchical subproblems. Our test set contains 65 main problems. We compute the Main score over these 65 problems and the Sub score over all 291 subtasks in the test set. Unlike standard coding tasks, it requires models to demonstrate deep domain knowledge, advanced reasoning, and multi-step scientific code synthesis.

\item\textbf{GPQA} \cite{rein2024gpqa} is a challenging question-answering benchmark containing high-quality multiple-choice questions across biology, physics, and chemistry. We utilize the {GPQA-Diamond} subset to evaluate the model's capability to synthesize complex scientific knowledge where simple information retrieval fails.
\end{itemize}
To strictly evaluate these diverse capabilities, we assign distinct agent pools based on task requirements. For text-centric reasoning tasks (AIME 2025, SciCode, and GPQA-Diamond), we employ the {Qwen series}, specifically Qwen-8B, Qwen-14B, and the Qwen3-30B-A3B variants (Thinking/Instruct).

\subsection{Baselines}

\begin{itemize}[leftmargin=*, nosep, itemsep=2pt, parsep=0pt, partopsep=0pt]
\item\textbf{MasRouter} \cite{yue-etal-2025-masrouter} addresses Multi-Agent System Routing with a cascaded controller network that determines the collaboration mode, progressively allocates agent roles, and assigns an LLM backbone to each agent. It thereby adapts model usage to each query while balancing task performance and inference overhead.

\item\textbf{AFlow}~\cite{zhang2025aflow} automatically searches over executable LLM workflows represented as directed graphs. Its optimizer iteratively revises workflow code using execution feedback; in our experiments, we use it as the topology-optimization baseline with a sample size of 3 and at most 5 optimization rounds.

\item\textbf{GraphPlanner}~\cite{feng2026graphplanner} is a graph-memory-augmented agentic router that jointly selects an agent role (Planner, Executor, or Summarizer) and an LLM backbone at each step, thereby producing a query-specific multi-agent workflow. It formulates this sequential construction process as an MDP and trains the policy with reinforcement learning, while its heterogeneous GARNet memory encodes query--agent--response interactions and historical workflows to support both inductive and transductive routing.

\item\textbf{OAgents}~\cite{zhu-etal-2025-oagents} is a general-purpose, modular agent framework derived from a controlled empirical study of agent design choices. It combines effective components for planning, acting, tool use, memory, and multi-agent collaboration, and serves as our representative full MAS baseline rather than a specialized router or topology optimizer.

\end{itemize}

\section{LLM Profile Construction}
\label{app:metrics}
\label{app:agent_profile}
Standard scalar metrics obscure the nuanced trade-offs essential for Multi-Agent Systems (MAS). To enable precise routing, we implement a multi-dimensional weighted evaluation framework, characterizing each LLM via five orthogonal metrics. This quantized metadata equips the optimizer with a high-fidelity capability landscape, facilitating optimal model-task alignment. To ensure reproducibility and representative performance, we aggregate foundational metrics from verified third-party evaluators, specifically Artificial Analysis \cite{artificial_analysis_2025} and OpenRouter \cite{OpenRouter}.

\subsection{Intrinsic Cognitive Capabilities}

This category quantifies a model’s cognitive upper bound, specifically its capacity to navigate novel, out-of-distribution (OOD) tasks without succumbing to hallucination. To operationalize this dimension, we introduce the \textbf{Frontier Reasoning Index (FRI)} While conventional evaluations often treat reasoning depth and hallucination rates as orthogonal metrics, we posit that in frontier domains, hallucination is a direct byproduct of insufficient reasoning. The FRI unifies these aspects, certifying a model’s ability to generate grounded deductions rather than stochastic fabrications when facing expert-level proofs and unseen research challenges.
We aggregate performance from three high-signal benchmarks: \textit{AA-Omniscience} \cite{artificialanalysis2025omniscience} (General Knowledge), \textit{GPQA Diamond} \cite{rein2024gpqa} (Expert-level Hallucination Resistance), and \textit{CritPt} \cite{zhu2025probing} (Frontier Physics). The index is formally defined as:

\begin{equation}
S_{\text{FRI}} = \alpha \cdot \bar{S}_{\text{Omni}} + \beta \cdot \bar{S}_{\text{GPQA}} + \gamma \cdot \bar{S}_{\text{Crit}}
\end{equation}

where $\bar{S}_{x}$ denotes the normalized score on dataset $x$. The coefficients $\alpha, \beta$, and $\gamma$ represent weighting factors for general knowledge, expert reasoning, and frontier exploration, respectively, calibrated such that $\alpha + \beta + \gamma = 1$. In our experiments, these are set to $0.4$, $0.3$, and $0.3$, respectively.
\subsection{Agentic Interaction Competencies}

Beyond static reasoning, effective agents require dynamic interaction capabilities. We characterize this dimension through three specialized metrics:
\begin{itemize}
    \item \textbf{Tool-Calling Precision (TCP).} TCP evaluates instrumental efficacy across diverse modalities (API, Python, Search) via $\tau^2$-Bench \cite{barres2025tau2}, targeting the semantic and temporal precision of tool invocation. We define $S_{\text{TCP}} = \mathcal{N}_{\tau}(s_{\text{raw}})$, normalizing tool selection and argument accuracy against the benchmark's optimal bounds.    \item \textbf{Structural Compliance Index (SCI).} Critical for MAS interoperability, SCI measures syntactic fidelity to rigid constraints (e.g., JSON) within \textit{IFBench} \cite{pyatkin2025generalizing}. We calculate $S_{\text{SCI}} = \mathcal{N}_{\text{fmt}}(s_{\text{raw}})$ to index the reliability of inter-agent data serialization, ensuring workflow stability against parsing failures.
    \item \textbf{Self-Remediation Index(SRI).} Serving as a proxy for execution survival rate, SRI quantifies the autonomy to recover from runtime exceptions by synthesizing alternative trajectories from error logs. We define $S_{\text{SRI}} = \mathcal{N}_{\text{hard}}(s_{\text{raw}})$, normalizing raw performance against the empirical ceiling of the \textit{Terminal-Bench} leaderboard \cite{merrill2026terminalbenchbenchmarkingagentshard}.
\end{itemize}

\subsection{Computational Utility}

Standard metrics focusing solely on verbosity or speed fail to reconcile diverse operational profiles: high-throughput generation, high-agility interaction, and the inherent latency characteristic of reasoning-intensive models. We therefore formulate \textbf{Adaptive Response Efficiency (ARE).} as a Max-Pooling Utility Function.This design permits utility maximization via either the throughput or agility pathway, while enforcing a hard Quality of Service (QoS) constraint to penalize the excessive latency often inherent in deep reasoning.The score is the maximum of two efficiency pathways, penalized by a QoS constraint:
\begin{equation}
\begin{aligned}
S_{\text{ARE}}(m) = &\max \left( E_{\text{thm}}(m), E_{\text{agi}}(m) \right) \\
&\cdot \mathcal{P}_{\text{qos}}(m).
\end{aligned}
\end{equation}

\textit{(1) Throughput Pathway ($E_{\text{thm}}$):} For generation-heavy tasks, we prioritize parallel decoding speed. We apply logarithmic normalization to smooth order-of-magnitude differences in TPS:
\begin{equation}
E_{\text{thm}}(m) = \frac{\ln(1 + \text{TPS}_m)}{\ln(1 + \text{TPS}_{\max})}.
\end{equation}

\textit{(2) Agility Pathway ($E_{\text{agi}}$):} For interactive, short-turn steps, we define effective information density. This pathway favors models with low Time-To-First-Token (TTFT) and conciseness (low Verbosity):
\begin{equation}
E_{\text{agi}}(m) = \frac{\kappa}{\sqrt{\text{TTFT}_m \cdot \text{Verbosity}_m}},
\end{equation}
where $\kappa$ is a scaling factor. This metric highlights models that deliver concise answers instantly, even if their generation speed is moderate.

\textit{(3) QoS Constraint ($\mathcal{P}_{\text{qos}}$):} To guarantee real-time responsiveness, we impose a soft latency barrier. If a model's TTFT exceeds a threshold $\tau$ (e.g., 10s), its utility decays exponentially:
\begin{equation}
\mathcal{P}_{\text{qos}}(m) =  \mathbb{I}(\delta_m \le 0) + \mathbb{I}(\delta_m > 0) \cdot e^{-\lambda \delta_m},
\end{equation}
where $\delta_m = \text{TTFT}_m - \tau$, ensuring that high-latency models are deprioritized regardless of their reasoning scores.

For completeness, we report the capability profiles used in our Qwen
experiments. For each dimension $j$, the raw score $x_{m,j}$ of model $m$ is
standardized over the four-model candidate pool and transformed as
\begin{equation}
\begin{aligned}
z_{m,j} &=
\frac{x_{m,j}-\mu_j}{\sigma_j},\\
\widetilde{S}_{m,j} &=
1+\frac{9}{1+\exp\!\left(-z_{m,j}/T\right)},
\qquad T=0.5.
\end{aligned}
\end{equation}
Here, $\mu_j$ and $\sigma_j$ denote the pool-wise mean and standard deviation.
If $\sigma_j<10^{-6}$, we assign the neutral score $5.0$ to all candidates.
The resulting values lie in $[1,10]$ and represent relative strengths within
the deployed pool rather than absolute benchmark performance.
The normalized profiles computed for the four Qwen models are summarized in
Table~\ref{tab:qwen_profiles}.

\begin{table}[h]
\centering
\scriptsize
\setlength{\tabcolsep}{3pt}
\resizebox{\columnwidth}{!}{\begin{tabular}{lccccc}
\toprule
\textbf{Model}
& \multicolumn{1}{c}{\textbf{Intrinsic}}
& \multicolumn{3}{c}{\textbf{Agentic Interaction}}
& \multicolumn{1}{c}{\textbf{Utility}} \\
\cmidrule(lr){2-2}\cmidrule(lr){3-5}\cmidrule(lr){6-6}
& \textbf{FRI} & \textbf{TCP} & \textbf{SCI} & \textbf{SRI} & \textbf{ARE} \\
\midrule
Qwen3-8B                    & 1.8 & 6.2 & 3.4 & 1.3 & 8.2 \\
Qwen3-14B                   & 2.5 & 8.9 & 2.1 & 6.9 & 7.0 \\
Qwen3-30B-A3B-Instruct-2507 & 8.4 & 1.3 & 5.0 & 8.9 & 8.2 \\
Qwen3-30B-A3B-Thinking-2507 & 9.3 & 7.6 & 9.7 & 6.9 & 1.3 \\
\bottomrule
\end{tabular}}
\caption{\textbf{Normalized capability profiles of the four Qwen backbones.}
Scores are relative to the deployed model pool; higher is better.}
\label{tab:qwen_profiles}
\end{table}

\section{Supplementary Experiments}
\label{app:supplementary}

\paragraph{Evaluator Backbone Sensitivity.}
\label{app:evaluator_sensitivity}

We investigate the sensitivity of MASA to the choice of evaluator backbone on AIME25. As shown in Table~\ref{tab:evaluator_sensitivity}, stronger evaluators generally improve Avg@3, while all three backbones achieve 100\% Pass@3, indicating that every problem is solved at least once across three samples. DeepSeek-V4-Flash attains the highest Avg@3 with latency comparable to Qwen-Flash, whereas GPT-OSS-120B incurs substantially higher planning overhead, demonstrating a clear accuracy--efficiency trade-off across evaluator choices.

\begin{table}[h]
\centering
\scriptsize
\setlength{\tabcolsep}{3pt}
\resizebox{\columnwidth}{!}{\begin{tabular}{lccccc}
\toprule
\textbf{Evaluator} &
\textbf{Avg@3} &
\shortstack{\textbf{Mean cost}\\\textbf{(\$/run)}} &
\shortstack{\textbf{Mean plan cost}\\\textbf{(\$/run)}} &
\textbf{Latency} &
\textbf{Plan Lat.} \\
\midrule
Qwen-Flash
& 90.28\% & \$0.15 & \$0.03 & 44.76s & 14.34s \\
GPT-OSS-120B
& 93.06\% & \$0.19 & \$0.05 & 59.14s & 28.27s \\
DeepSeek-V4-Flash
& \textbf{97.22\%} & \$0.20 & \$0.05 & 44.68s & 16.18s \\
\bottomrule
\end{tabular}}
\caption{\textbf{Evaluator sensitivity on AIME25.} Each cost column reports the mean per run across the three runs used to compute Avg@3. Stronger evaluators improve Avg@3 with moderate planning overhead, while DeepSeek-V4-Flash achieves the best accuracy–latency trade-off.}
\label{tab:evaluator_sensitivity}
\end{table}

\paragraph{Predictive Evaluation versus Execution-Score Evaluation.}
\label{app:predictive_vs_execution}

As a supplementary control, we use test-set outcomes to construct an execution-score-only alternative to our multidimensional predictive evaluator. We use these outcomes only for this control. The main comparison in Table~\ref{tab:main_results} evaluates \textsc{Masa} and the reproduced baselines without access to test labels. As shown in Table~\ref{tab:predictive_vs_execution}, replacing the profile-based evaluator with a single scalar execution score reduces AIME Pass@3 by 33.33 points, while increasing total cost by 2.22$\times$ and latency by 2.73$\times$. This result suggests that fine-grained modeling of reasoning ability, tool use, reliability, cost, and latency provides substantially more informative routing signals than aggregate execution performance alone.

\begin{table}[h]
\centering
\small
\setlength{\tabcolsep}{3pt}
\resizebox{\columnwidth}{!}{\begin{tabular}{lccc}
\toprule
\textbf{Evaluator} &
\textbf{Pass@3} &
\textbf{Total cost} &
\textbf{Latency} \\
\midrule
\textbf{Multidimensional predictor}
& \textbf{100.00\%} & \textbf{\$0.44} & \textbf{44.76s} \\
Execution score only
& 66.67\% & \$0.97 & 122.20s \\
\bottomrule
\end{tabular}}
\caption{\textbf{Supplementary evaluator control on AIME25.} Total cost sums all three runs. The cost ratio uses the unrounded totals. Multidimensional capability modeling outperforms an execution-score-only evaluator in accuracy, cost, and latency.}
\label{tab:predictive_vs_execution}
\end{table}

\end{document}